\documentclass{article}
\usepackage{spconf}

\usepackage{enumitem}
\usepackage{nicefrac}
\usepackage{microtype}

\usepackage{graphicx}
\usepackage{subcaption}

\usepackage{booktabs}
\usepackage{threeparttable}
\usepackage{multirow}
\usepackage{adjustbox}

\usepackage{amsfonts,amsmath,amssymb}
\usepackage{algorithm,algorithmicx,algpseudocode}

\usepackage{hyperref}
\usepackage{url}

\usepackage{color}
\definecolor{white}{RGB}{200,200,200}
\definecolor{black}{RGB}{40,44,52}
\definecolor{darkred}{rgb}{0.8,0,0}
\definecolor{darkgreen}{rgb}{0,0.7,0}
\definecolor{darkblue}{rgb}{0,0.0,0.3}
\hypersetup{
  colorlinks,
  linkcolor=darkred,
  citecolor=darkgreen,
  urlcolor=darkblue,
  pdfinfo={
    Title={Continual Local Training for Better Initialization of Federated Models},
    Author={Xin Yao, Lifeng Sun},
  }
}

\title{Continual Local Training for Better Initialization of \\Federated Models}
\name{Xin Yao\textsuperscript{\rm 1,2}, Lifeng Sun\textsuperscript{\rm 1,2}\thanks{© 2020 IEEE. Personal use of this material is permitted. Permission from IEEE must be obtained for all other uses, in any current or future media, including reprinting/republishing this material for advertising or promotional purposes, creating new collective works, for resale or redistribution to servers or lists, or reuse of any copyrighted component of this work in other works.}}
\address{
  \textsuperscript{\rm 1}BNRist, Department of Computer Science and Technology, Tsinghua University\\
  \textsuperscript{\rm 2}Key Laboratory of Pervasive Computing, Ministry of Education\\
  \texttt{yaox16@mails.tsinghua.edu.cn, sunlf@tsinghua.edu.cn}
}
\begin{document}
%
\maketitle
\begin{abstract}
Federated learning (FL) refers to the learning paradigm that trains machine learning models directly in the decentralized systems consisting of smart edge devices without transmitting the raw data, which avoids the heavy communication costs and privacy concerns.
Given the typical heterogeneous data distributions in such situations, the popular FL algorithm \emph{Federated Averaging} (FedAvg) suffers from weight divergence and thus cannot achieve a competitive performance for the global model (denoted as the \emph{initial performance} in FL) compared to centralized methods.
In this paper, we propose the local continual training strategy to address this problem.
Importance weights are evaluated on a small proxy dataset on the central server and then used to constrain the local training.
With this additional term, we alleviate the weight divergence and continually integrate the knowledge on different local clients into the global model, which ensures a better generalization ability.
Experiments on various FL settings demonstrate that our method significantly improves the initial performance of federated models with few extra communication costs.
\end{abstract}
\section{Introduction}
\label{sec:intro}

Recent years have witnessed a rapid growth in the application of smartphones, wearable devices, and Internet of Things (IoT) devices.
These smart devices and sensors are generating a huge amount of valuable yet underutilized data.
Unfortunately, the privacy risks would prohibit transmitting the raw data to the data center for further use while the limited networking availability and capacity of edge devices could not afford training through a standard distributed optimization strategy.

In the presence of user-generated data distributed among a huge amount of edge devices (or \texttt{clients}) in non-IID manners, Federated Learning (FL) \cite{konevcny2015federated,mcmahan2017communication} proposes a communication-efficient algorithm to train machine learning models without transmitting the raw data.
A typical FL system (or \texttt{server}) selects a part of available devices and sends them the global model as the initialization.
Then the selected clients compute an update to the global model with their local data using some pre-designed optimization methods.
Finally the server collects all the updates and aggregates them to get the new global model.
Such an iteration is called a \emph{round} in FL optimization and repeated iteratively until convergence.

\noindent\textbf{Challenges in Federated Learning}

There are mainly three challenges that distinguish FL from traditional distributed learning problems.

The first one is the system challenge, i.e., a massive number of edge clients with limited network connections.
To cope with this, Federated Averaging (FedAvg) \cite{mcmahan2017communication} proposes selecting a subset of clients for participating in training at each round.
And Yao et al.~\cite{yao2018twostream,yao2019towards} introduce additional mechanisms to local training procedures to accelerate the convergence and thus reduce the overall communication costs.

The second one is the privacy preservation, e.g., differential privacy guarantees \cite{agarwal2018cpsgd}, homomorphic encryption \cite{bonawitz2017practical}, and secure frameworks \cite{liu2018secure}.
These researches are relatively independent and can be integrated into most FL frameworks.

The last one is the statistic challenge, i.e., the heterogeneous or non-IID data distributions on clients, which is also the main focus of this paper.
The most relevant research work to ours are two papers as follows.
Zhao et al. \cite{zhao2018federated} demonstrate that the accuracy reduction, especially when the local data are distributed in a non-IID manner, can be attributed to the weight divergence between the local and global models.
They further propose a data-sharing strategy to alleviate this problem by creating a small subset of data (usually called the \emph{proxy dataset}) that is shared among all the clients.
Similar strategies have also been adopted in \cite{jeong2018communication,yao2019fedmeta}.
FedCurv \cite{shoham2019overcoming}, inspired by Elastic Weight Consolidation \cite{kirkpatrick2017overcoming} (EWC, a typical continual learning algorithm), calculates the Fisher information matrix of the local models, and use them to reduce the weight divergence during local training.
However, even with bandwidth optimization tricks, the amount of parameters to be transmitted in FedCurv is three times as many as that in FedAvg.

\noindent\textbf{Continual Learning}

\emph{Continual learning}, or \emph{lifelong learning}, is a research field in transfer learning that tries to avoid the \emph{catastrophic forgetting} \cite{goodfellow2013empirical} when learning task \textit{B} based on the model trained on task \textit{A}, or in other words, to enable the well-trained model to learn a new task without forgetting previously learned tasks.
The key idea of continual learning algorithms is to find a common feature space \cite{li2017learning,yao2019adversarial} or parameter space \cite{kirkpatrick2017overcoming,aljundi2018memory} for the old and new tasks so that the model can work well on the new task without forgetting old tasks.

Due to the heterogeneous data distributions in FL, the local models on clients are trained with only their local data without the information of the global data distribution.
This is exactly the weight divergence as mentioned before, which can be alleviated by finding a common feature space or parameter space for the local and global models.

In this paper, we propose federated learning with local continual training (FedCL, CL for Continual Learning) strategy to alleviate the weight divergence and continually integrate the knowledge on different local clients into the global model, which ensures a better generalization ability for the global model.
Importance weights w.r.t. the global model parameters are evaluated on a small proxy dataset on the central server and then used to constrain the local training.
In this way, the federated model is able to learn knowledge of clients while keeping its original performance.
Furthermore, with bandwidth optimization tricks, the extra communication costs can be as low as 5\% of the total ones.

In conclusion, our contributions are as follows:
\begin{itemize}
    \item We propose a continual local training strategy with the globally estimated importance weights to improve the generalization ability of federated models.
    \item Experiments on popular FL settings show that the proposed method can greatly improve the initial performance of federated models with as low as 5\% extra communication costs.
\end{itemize}

\begin{figure*}[t]
  \begin{subfigure}[b]{.245\linewidth}
    \centering
    \includegraphics[width=\linewidth]{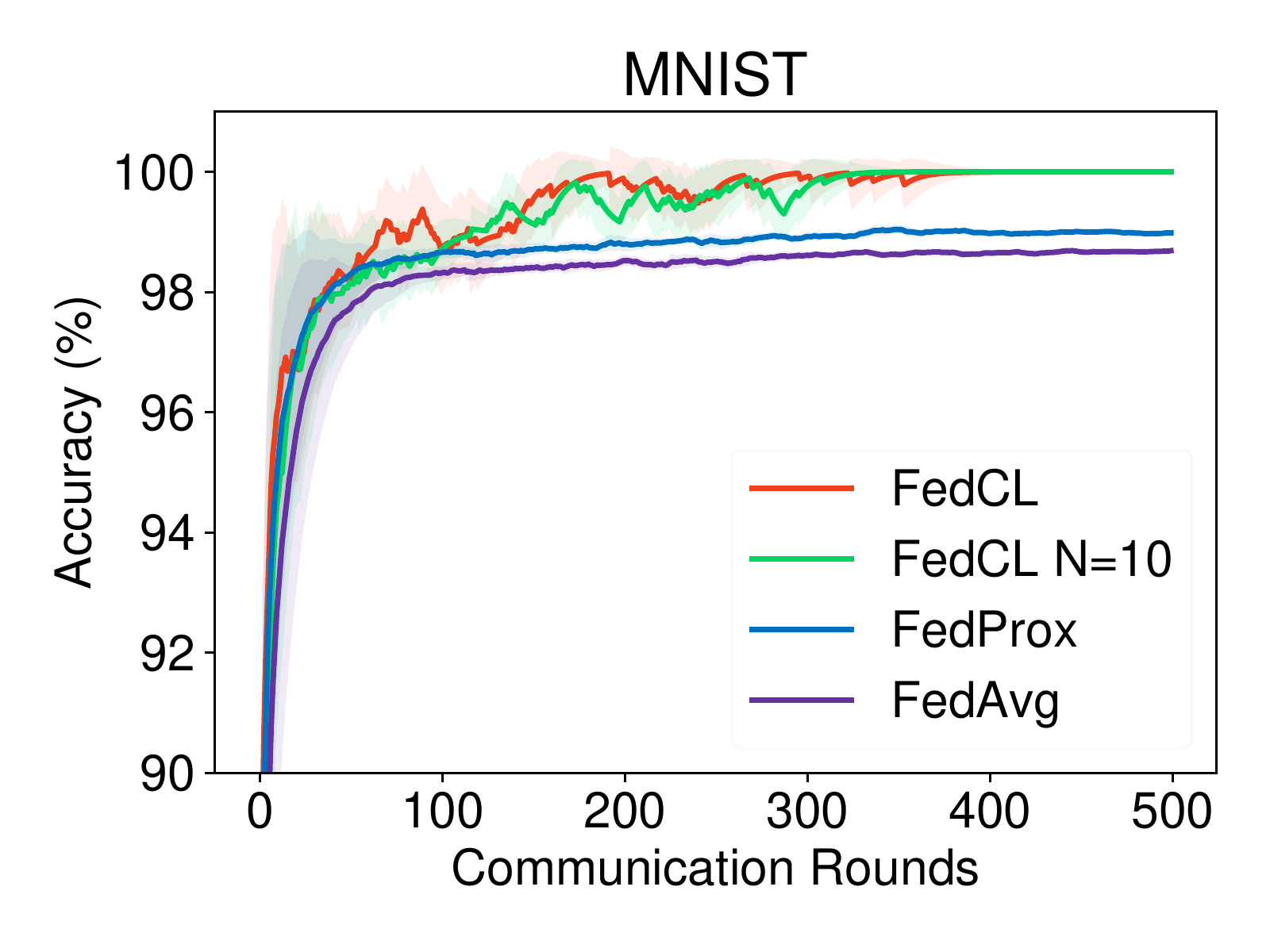}
    \includegraphics[width=\linewidth]{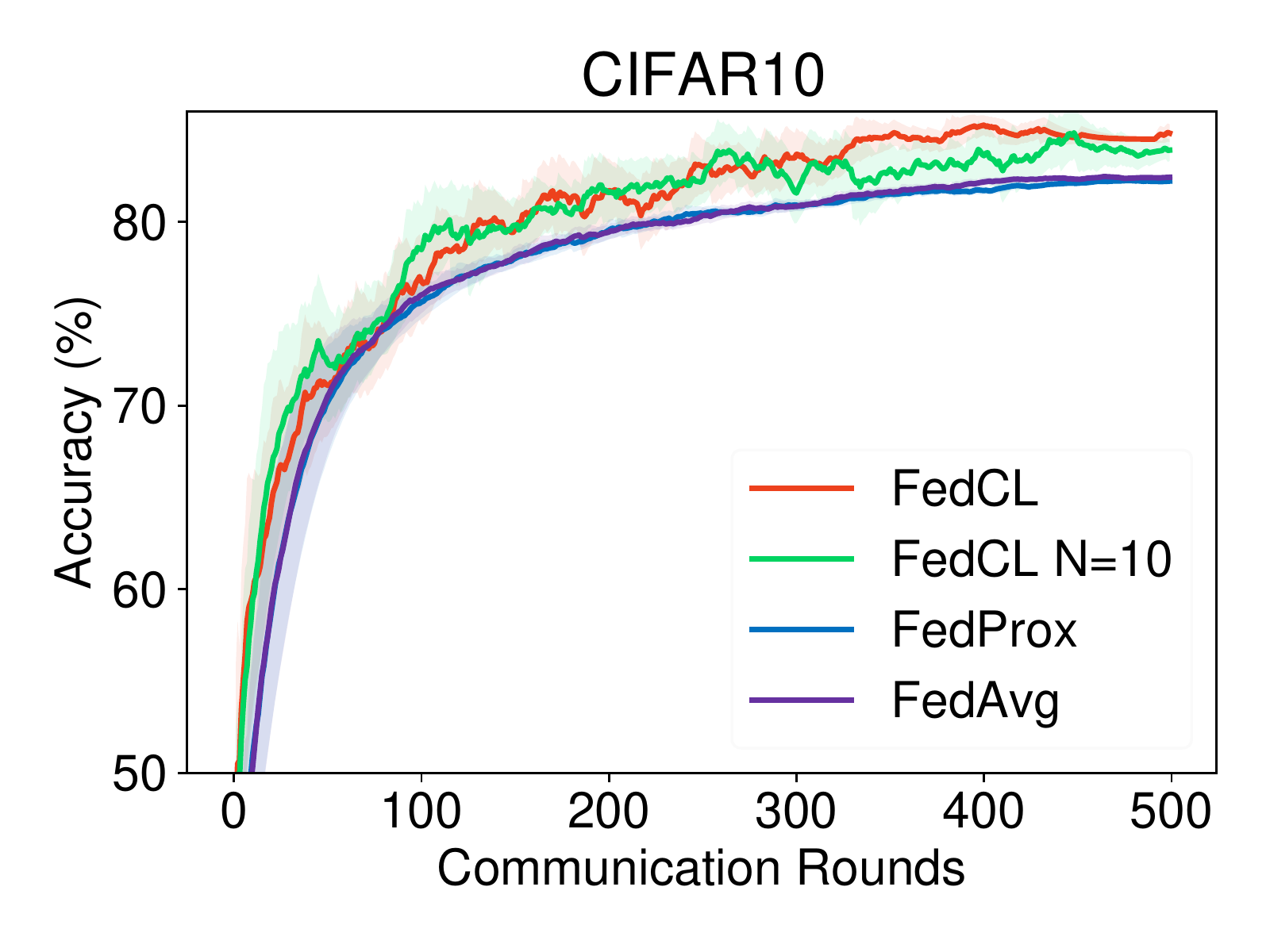}
    \caption{Uniform}
  \end{subfigure}
  \begin{subfigure}[b]{.245\linewidth}
    \centering
    \includegraphics[width=\linewidth]{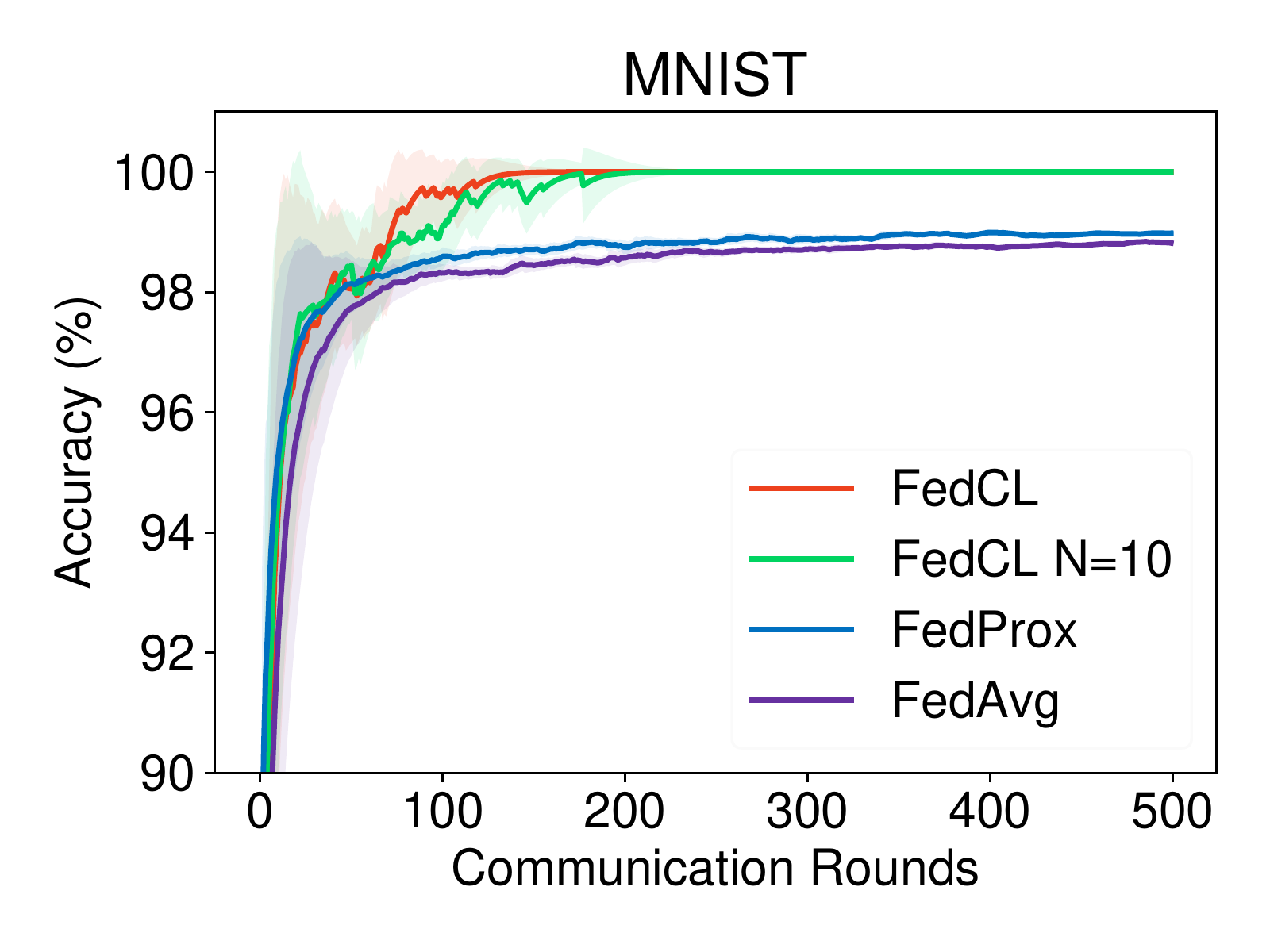}
    \includegraphics[width=\linewidth]{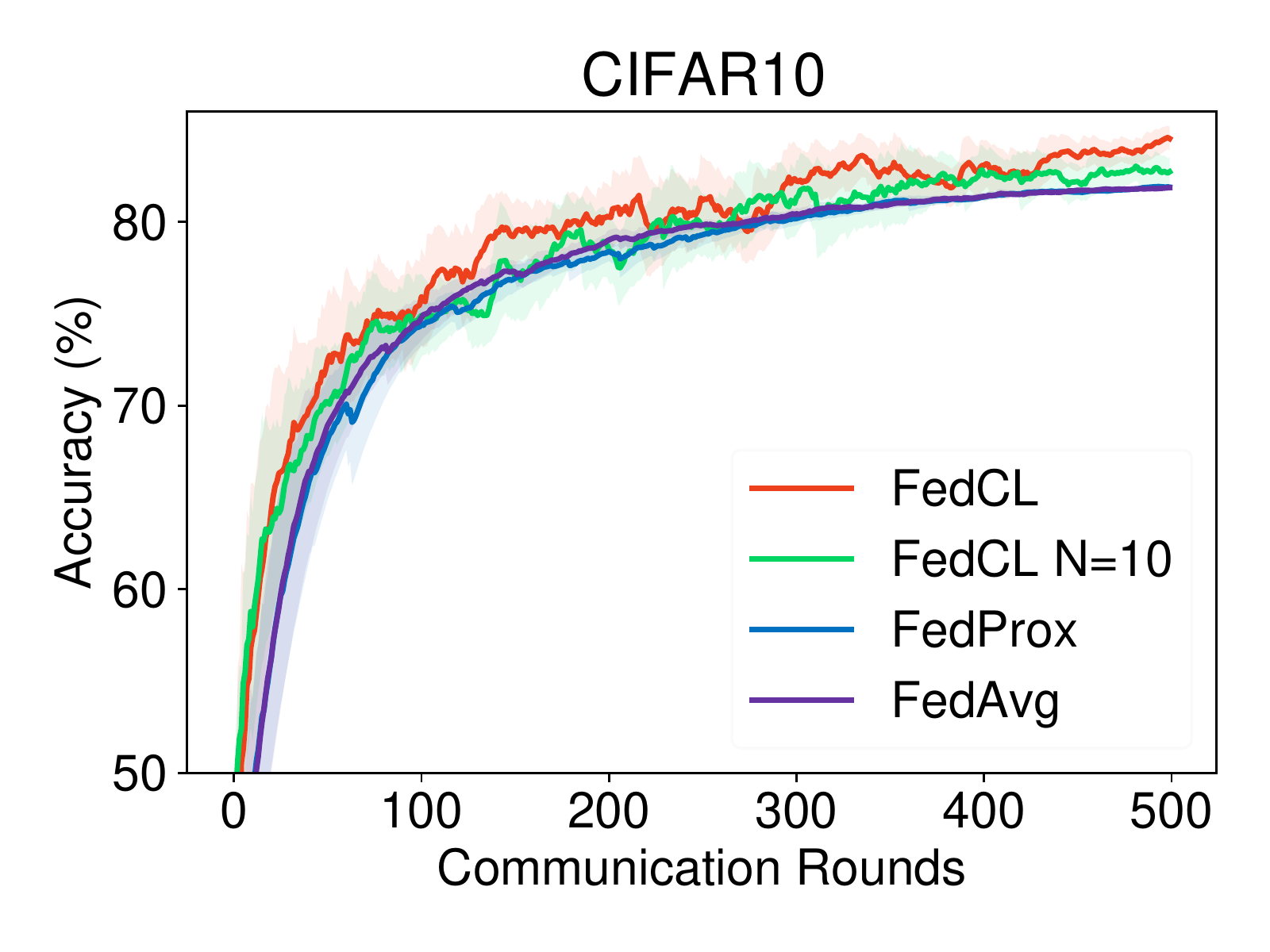}
    \caption{$\alpha = 100$}
  \end{subfigure}
  \begin{subfigure}[b]{.245\linewidth}
    \centering
    \includegraphics[width=\linewidth]{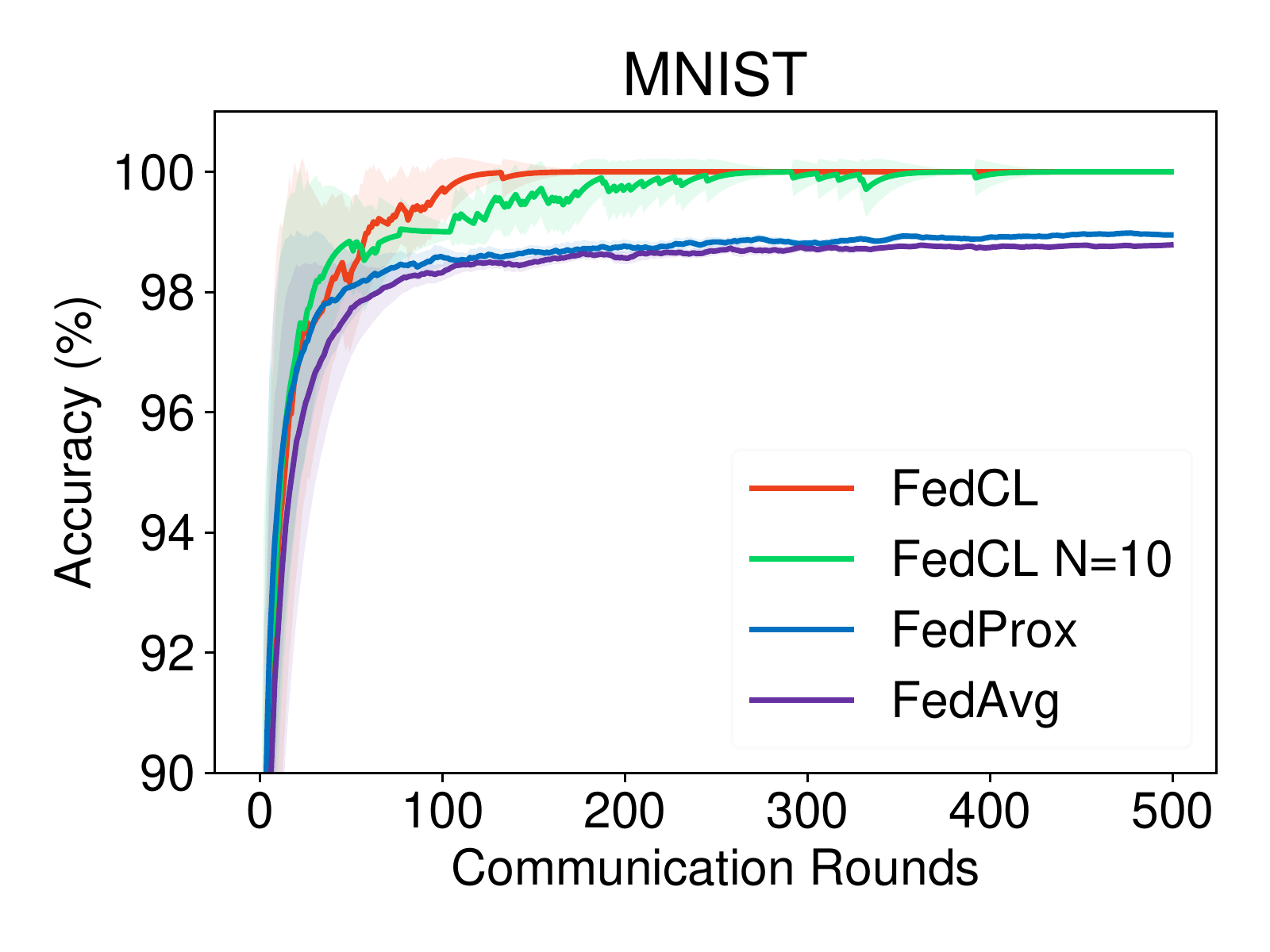}
    \includegraphics[width=\linewidth]{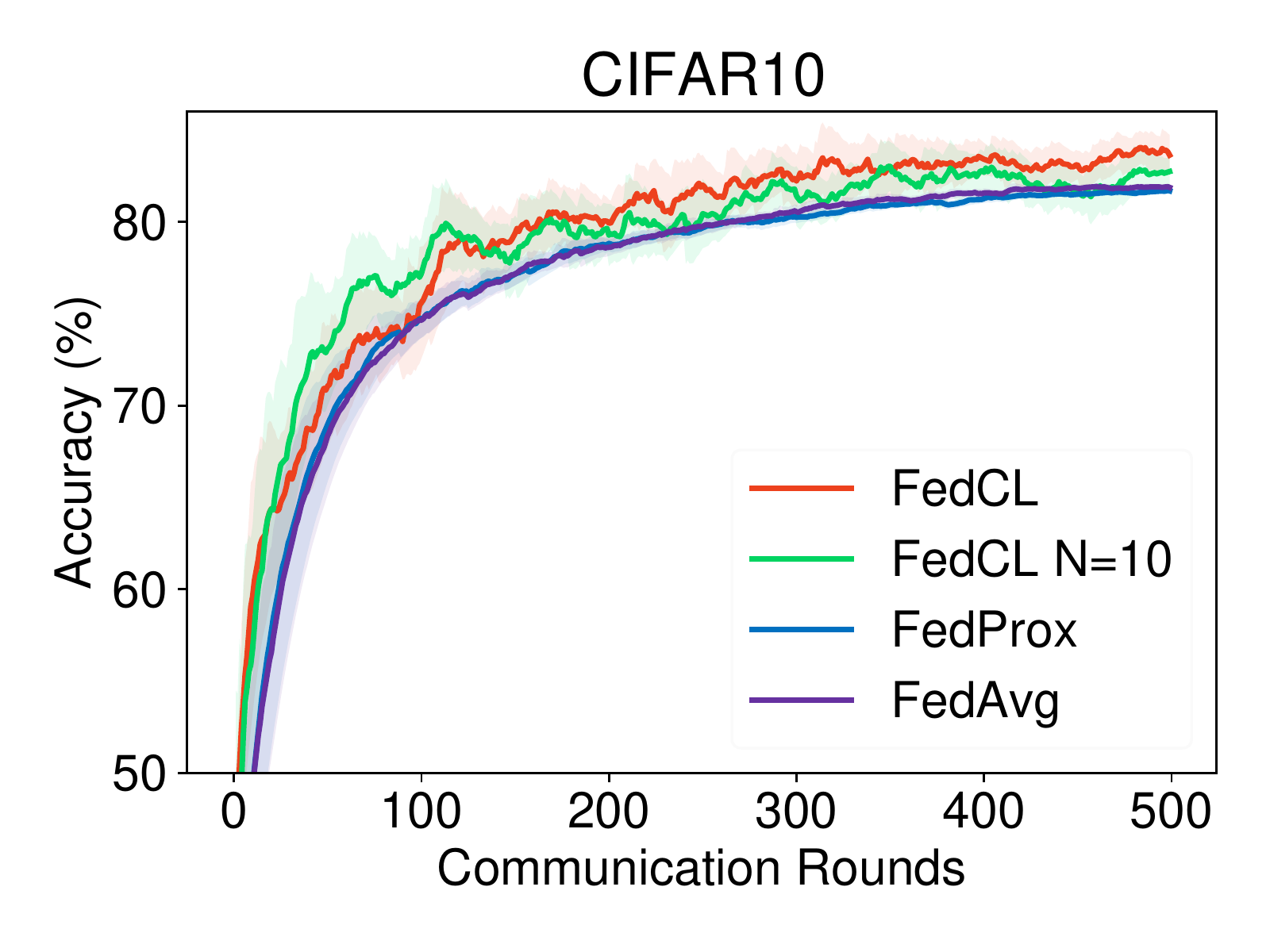}
    \caption{$\alpha = 10$}
  \end{subfigure}
  \begin{subfigure}[b]{.245\linewidth}
    \centering
    \includegraphics[width=\linewidth]{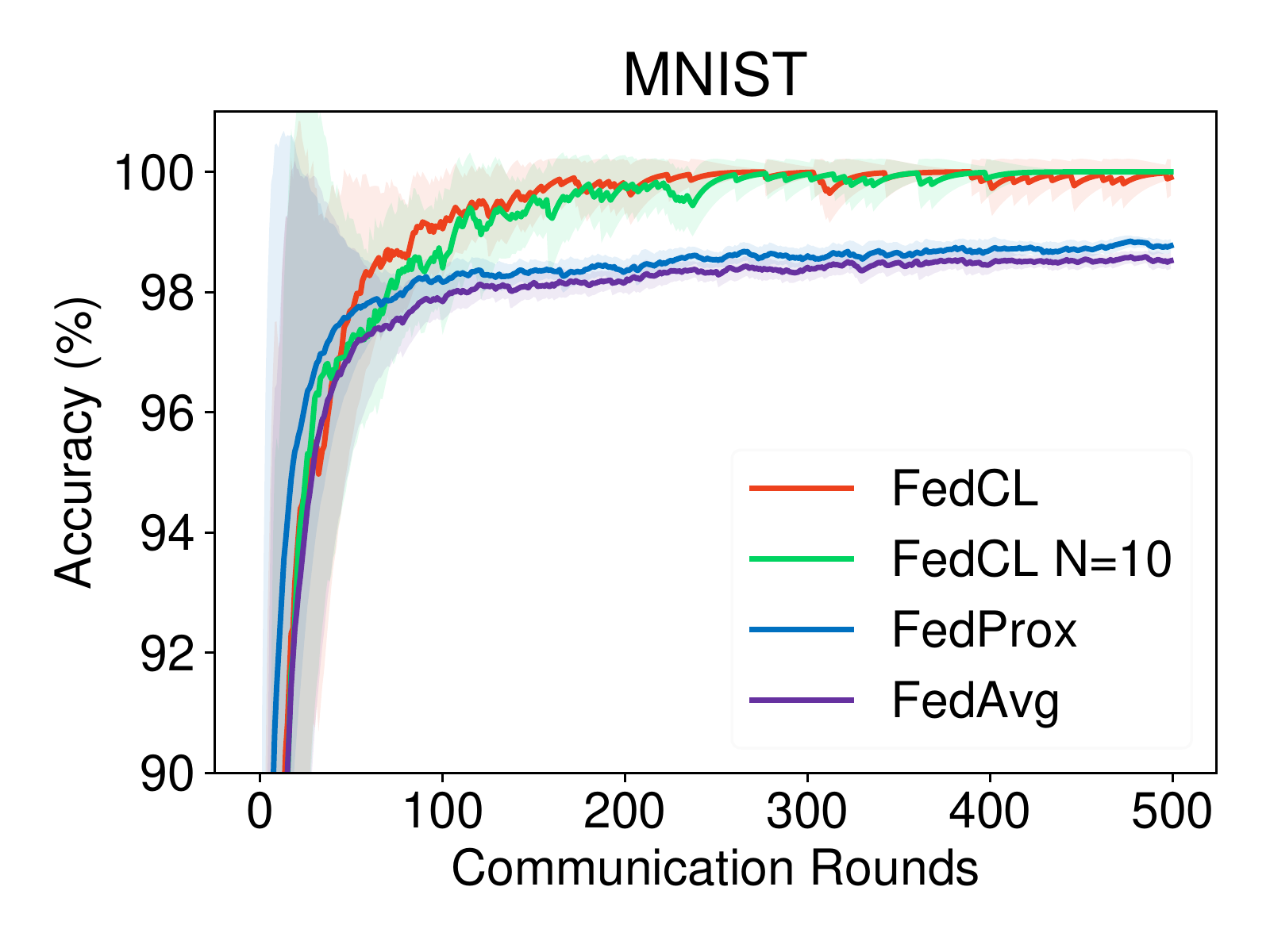}
    \includegraphics[width=\linewidth]{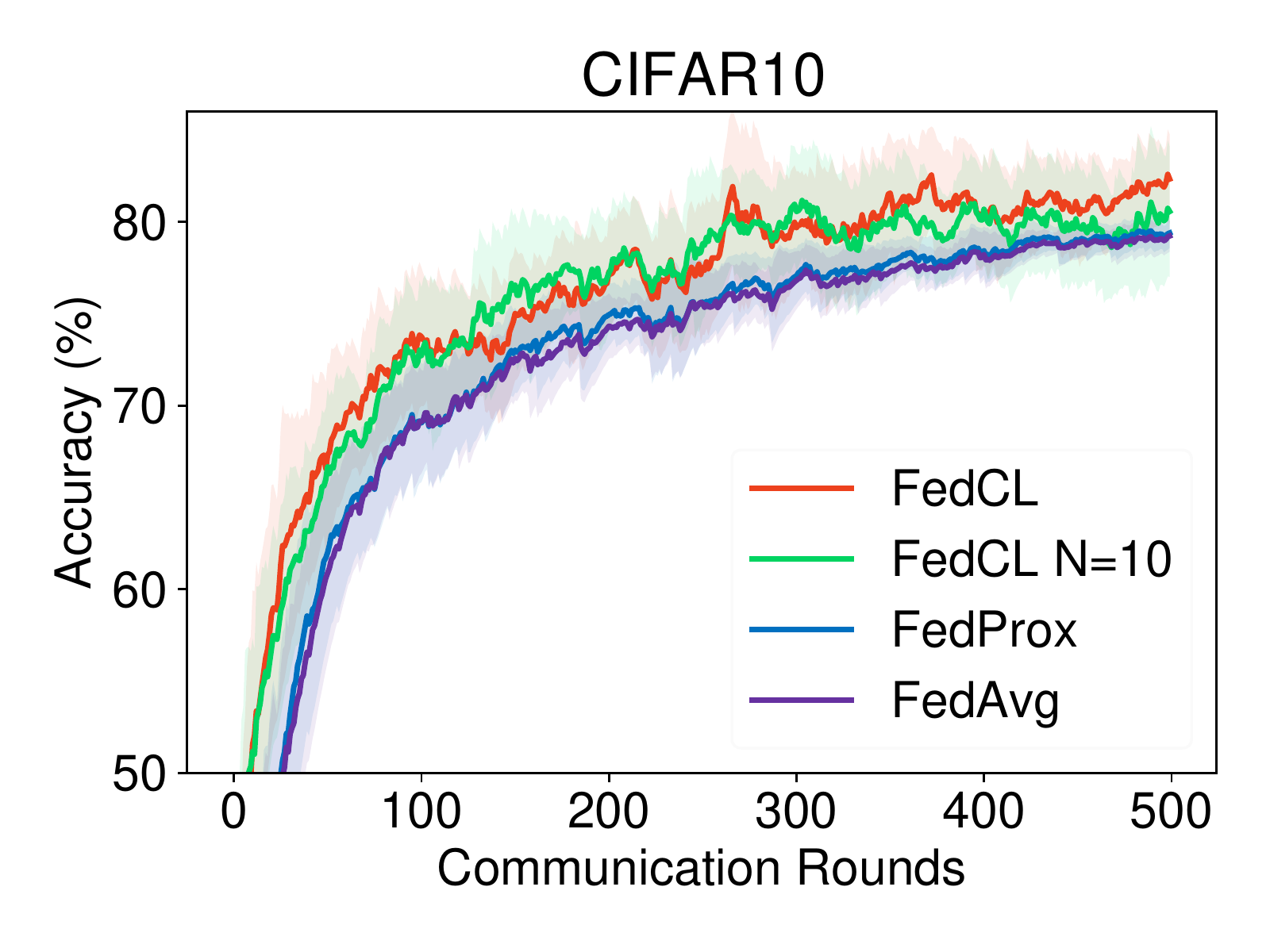}
    \caption{$\alpha = 1$}
  \end{subfigure}
\caption{The learning curves on MNIST and CIFAR10 with different local identicalness: Uniform ($\alpha \rightarrow +\infty$) and $\alpha = 100,10,1$, where $\alpha$ is the concentration parameter of Dirichlet distribution (Section \ref{sec:setup}). (Better viewed in color)}
\label{fig:results}
\end{figure*}

\section{Methods}
\label{sec:methods}

At a high level, FedCL follows the parameter-regularization\footnote{Refer to \cite{yao2019adversarial} for the categorizations of continual learning algorithms} continual training paradigm on clients, i.e., penalizing the important parameters of the global model for changing.
Instead of estimating the importance weight of model parameters on clients and exchanging them as in \cite{shoham2019overcoming}, which will bring at least twice the extra communication costs, FedCL estimates the importance weights on the proxy dataset on the server and then distributes them to the clients.
Even in the worst case, FedCL requires 1.5 times of the communication costs in total.
Further, we find that the importance weights do not change very frequently so that they can be estimated and distributed at an interval, e.g., ten rounds, which will further reduce the extra communication costs to 5\%.
It is worth noting that FedCL is compatible with most of the parameter-regularization continual learning algorithms, e.g., SI \cite{zenke2017continual} and MAS \cite{aljundi2018memory}.
In this paper, we adopt the Elastic Weight Consolidation (EWC) \cite{kirkpatrick2017overcoming}.

\begin{algorithm}[t]
  \caption{FL with Continual Local Training}
  \label{algorithm}
  \begin{algorithmic}[1]
    \Require (Interval $N$)
    \State initialize $\theta_0$
    \For {each round t = 0, 1, 2, ...}
      \State Estimate $\Omega$ with Eq.(\ref{eq:ewc_1})(\ref{eq:ewc_2})
      \State $m \leftarrow \text{max}(C \cdot K, 1)$
    	\State $S_t \leftarrow$ random sample $m$ clients
        \For {each client $k \in S_t$} in parallel
          \If{$t~\text{mod}~N == 0$}
            \State $\theta^k_{t+1} \leftarrow \textbf{ClientUpdate}(\theta_t, \Omega)$
          \Else
            \State $\theta^k_{t+1} \leftarrow \textbf{ClientUpdate}(\theta_t)$
          \EndIf
        \EndFor
        \State $\theta_{t+1} \leftarrow \frac{1}{\left\vert\mathcal{D}_{S_t}\right\vert}\sum_{k \in S_t} \left\vert\mathcal{D}_k\right\vert \theta^k_{t+1}$
    \EndFor
  \end{algorithmic}
  \begin{algorithmic}[1]
    \Ensure ($\theta_t$, $\Omega$)
    \If{$\Omega$ is not received}
      \State $\Omega \leftarrow I$
      \Comment{$I$ is the identity matrix}
    \EndIf
    \State Optimize Eq.(\ref{eq:loss}) with local data $\mathcal{D}_k$ for $E$ epochs to get updated local model $\theta^k_{t+1}$
    \State return $\theta^k_{t+1}$ to server
  \end{algorithmic}
\end{algorithm}

\subsection{Estimate Importance Matrix on Proxy Data}
\label{sec:method_center}

How to estimate the importance weight of model parameters is the key contribution of parameter-regularization continual learning algorithms but not the focus of this work.
EWC suggests estimating the importance weight matrix of the global model $\theta_g$ using the diagonal of its Fisher information matrix.
The justification is included in \cite{kirkpatrick2017overcoming}.

Specifically, in deep neural networks, the importance weights can be computed from first-order derivatives. Concretely, give a data sample $(x_k, y_k) \in \mathcal{D}_{proxy}$ with $\mathcal{D}_{proxy}$ the proxy dataset on the server, and global model parameter $\theta_g$, the importance weights can be approximated by:
\begin{equation}
\label{eq:ewc_1}
  \omega^k = \left\lVert\frac{\partial \mathcal{L}(\theta_g(x_k), y_k)}{\partial \theta_g}\right\rVert_2
\end{equation}
where $\mathcal{L}$ is the loss function and $\theta_g(x_k)$ is the output of the global model w.r.t. data sample $(x_k, y_k)$.
Then we accumulate $\omega^k$ over the proxy dataset to obtain the importance weight $\Omega$ for parameter $\theta_g$:
\begin{equation}
\label{eq:ewc_2}
  \Omega = \frac{1}{|\mathcal{D}_{proxy}|}\sum_{(x_k, y_k) \in \mathcal{D}_{proxy}} \omega^k
\end{equation}
where $|\mathcal{D}_{proxy}|$ is the number of examples in proxy dataset.
Here we get the importance weight matrix $\Omega$.

\subsection{Continual Training on Clients}
\label{sec:method_local}

The basic assumption of parameter-regularization continual learning algorithms is that deep neural networks are over-parameterized so that it is able to find an optimal solution $\theta^*_B$ for task \textit{B} that is close to the previously found solution for task \textit{A}, $\theta^*_A$.
In FL, it is to find an optimal solution $\theta^*$ for both the local and global data distributions while keeping the generalization ability of the global model $\theta^*_g$.

Formally, in the FL system containing $K$ clients in total, supposing that we have the importance weight matrix $\Omega = \{\Omega_{ij}\}$ estimated in Section \ref{sec:method_center}, we optimize the following empirical risk on client $k$ for $k \in C \cdot K$ ($C$ is the fraction of selected clients):
\begin{equation}
\label{eq:loss}
  \mathcal{L}_k(\theta_k) = \mathcal{L}_{local}(\theta_k) + \lambda \sum_{i,j}\Omega_{ij} (\theta_{k,ij} - \theta_{g,ij})^2
\end{equation}
where $\mathcal{L}_{local}$ is the original loss function on local client $k$ and $\lambda$ is a hyper parameter.
With the loss function Eq.(\ref{eq:loss}), we force the model $\theta_k$ to fit the local data distribution while keeping the knowledge of the global model $\theta_g$, which alleviates the weight divergence between the local and global model and thus ensures a better generalization and initial accuracy.

\subsection{Cut Down Bandwidth}

Following the procedures in Section \ref{sec:method_center} and \ref{sec:method_local}, the FL system consumes 50\% extra communication costs in each round (the same parameters to be transmitted from clients to server while double ones from server to clients).

Considering that the global model parameters do not change violently \cite{zhao2018federated}, we propose estimating and distributing the importance weight matrix $\Omega$ at an interval of $N$ rounds.
Through experiments, we show that when $N = 10$, indicating only 5\% extra communication costs, the local continual training strategy takes effect on alleviating the weight divergence as well.

The whole procedures of the proposed FedCL are organized as Algorithm \ref{algorithm}.

\section{Experiments}
\label{sec:exp}


\subsection{Experimental Setup}
\label{sec:setup}

\textbf{Datasets}

We use MNIST \cite{lecun1998gradient} and CIFAR10 \cite{krizhevsky2009learning} as basic datasets in our experiments.
To simulate the FL settings, especially the non-IID data distribution on clients, we follow a recent research work \cite{hsu2019measuring} to synthesize the non-IID data distributions with a continuous range of identicalness using the Dirichlet distribution.
Concretely, we assume that the class labels of examples on every clients follow a categorical distribution over $M$ classes parameterized by a vector $\mathbf{q}$ ($q_i \ge 0, i\in[1, M]$ and $\|\mathbf{q}\|_1 = 1$), where $\mathbf{q} \sim \texttt{Dirichlet}(\alpha\mathbf{I})$ with $\alpha > 0$ being the concentration parameter controlling the identicalness among clients.
A bigger $\alpha$ indicates a more uniform distribution.

We set up ten clients and set $C = 0.2$, i.e., selecting two clients for participating in training each round.
And we randomly sample 1\% of the total examples as the proxy dataset on the server for FedCL.

\noindent\textbf{Models}

For MNIST based settings, we use the same CNN architecture as FedAvg: two 5$\times$5 convolution layers (the first with 32 channels while the second with 64, each followed by a ReLU activation and 2$\times$2 max pooling), a fully connected layer with 512 units (with a ReLU activation and random dropout), and a final softmax output layer.

For CIFAR10 based settings, we use a CNN model with two 5$\times$5 convolution layers (both with 64 channels, each followed by a ReLU activation and 3$\times$3 max pooling with stride size 2), two fully connected layers (with 384 and 192 units respectively, each followed by a ReLU activation and random dropout) and a final softmax output layer.

We set $E = 2$ (local epochs) and $B = 64$ (batch size for local training), and use the SGD optimizer with the learning rate 0.005 and a decay rate 0.99 per communication rounds.
The hyper parameter $\lambda$ in Eq.(\ref{eq:loss}) is set to 0.5.
We compare the proposed FedCL with the vanilla FedAvg \cite{mcmahan2017communication} and FedProx \cite{li2018federated}.
Besides, in this paper, we focus on the average accuracy of the local model on clients, which is denoted as \emph{initial accuracy} and reflects the generalization ability of federated models.

\begin{figure}
  \begin{subfigure}[b]{.49\linewidth}
    \centering
    \includegraphics[width=\linewidth]{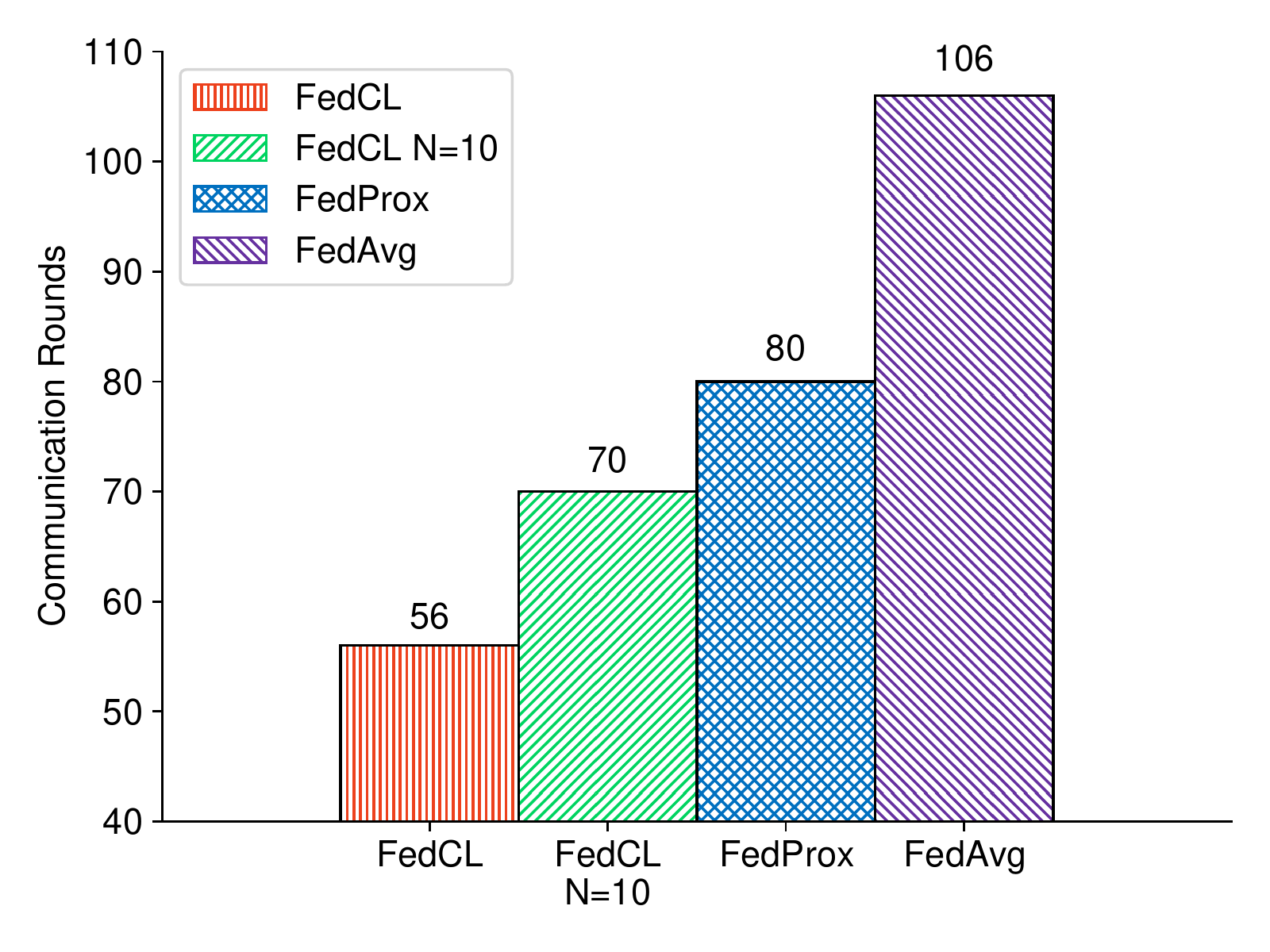}
    \caption{98\% on MNIST}
  \end{subfigure}
  \begin{subfigure}[b]{.49\linewidth}
    \centering
    \includegraphics[width=\linewidth]{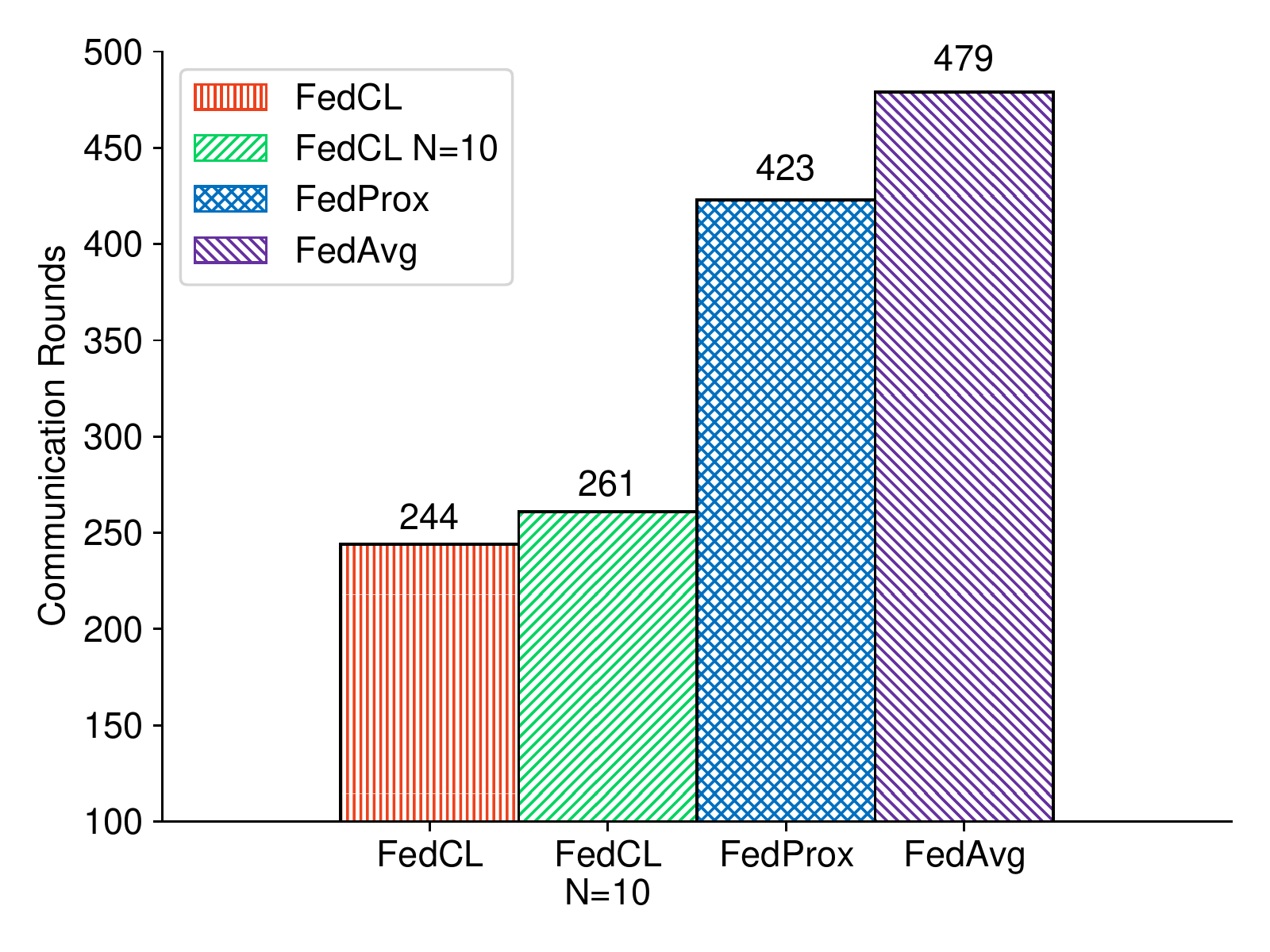}
    \caption{79\% on CIFAR10}
  \end{subfigure}
  \caption{Number of communication rounds to reach the accuracy of 98\% on MNIST and 79\% on CIFAR10 with local identicalness $\alpha = 1$. (Better viewed in color)}
  \label{fig:rounds}
\end{figure}

\subsection{Results and Analysis}
\label{sec:results}

We first report the learning curves, i.e., the test accuracy vs. communication rounds, on MNIST (upper row) and CIFAR10 (lower row) with different local identicalness $\alpha$ as Fig. \ref{fig:results}.
As we can see, on both datasets, FedCL greatly improves the initial accuracy of the global model compared to FedProx and the vanilla FedAvg.
It is worth noting that when the importance weight matrix $\Omega$ is estimated and distributed at an interval $N=10$, FedCL N=10 still reaches convergence faster and achieves better performance.
Considering different local data identicalness $\alpha$, FedCL outperforms FedProx and FedAvg under all the circumstances, especially when the local data distributions are extremely non-IID.
In such cases, the weight divergence is more critical and thus the performance drop is inevitable.
FedCL (with or without an interval) alleviates this problem by keeping the important parameters in the global model from changing during local training with the help of centrally estimated importance matrix $\Omega$.
In this way, the federated model continually integrates the knowledge on local clients while keeping its original performance.

We then illustrate the number of communication rounds to reach certain accuracies on MNIST and CIFAR10 with local identicalness $\alpha = 1$ as Fig. \ref{fig:rounds}.
Compared to FedAvg, FedCL requires only 52.8\% and 50.9\% of the communication rounds to reach the accuracy of 98\% on MNIST and 79\% on CIFAR10 respectively.
When the importance matrix $\Omega$ is estimated at an interval of ten rounds, indicating only 5\% extra communication costs, FedCL N=10 still reduces the needed communication rounds by up to 45.5\%.
These results show that the proposed local continual training strategy greatly accelerates the convergence rate of the global model in terms of initial accuracy.
We specifically demonstrate the results with $\alpha = 1$ because the non-IID situation can better prove the effectiveness of the proposed method.

Finally, we summarize the convergence initial accuracy and personalize accuracy (the average accuracy after local training) of all the compared methods on CIFAR10 with $\alpha=10$ and $1$ as Table \ref{tab:accuracy}.
With no surprise, FedCL (with or without an interval) greatly improves the initial accuracy while achieving very close personalize accuracy of federated models compared to FedProx and FedAvg.
According to this result, we conclude that the constraint used in local continual training strategy is just fine to keep the global knowledge of federated models while allowing it to fit the local data simultaneously.

\begin{table}
\caption{The convergence \emph{initial accuracy} and \emph{personalize accuracy} of compared methods on CIFAR10 with $\alpha=10,1$.}
\label{tab:accuracy}
\begin{tabular}{@{}lcccc@{}}
  \toprule
  \multirow{2}{*}{} & \multicolumn{2}{c}{$\alpha$ = 10}                                    & \multicolumn{2}{c}{$\alpha$ = 1}                                     \\ \cmidrule(l){2-3}\cmidrule(l){4-5} 
                    & \multicolumn{1}{l}{Initial} & \multicolumn{1}{l}{Personalize} & \multicolumn{1}{l}{Initial} & \multicolumn{1}{l}{Personalize} \\ \midrule
  FedAvg            & 81.93                       & 83.10                           & 79.23                       & 84.64                           \\
  FedProx           & 81.73                       & 83.33                           & 79.51                       & 85.35                           \\ \midrule
  FedCL             & \textbf{84.05}              & 83.08                           & \textbf{82.59}              & 84.95                           \\
  FedCL N=10        & 83.03                       & 82.96                           & 81.17                       & 84.82                           \\ \bottomrule
  \end{tabular}
\end{table}

\section{Conclusion}
\label{sec:conclusion}

In this paper, we propose a continual local training strategy, guided by the importance weight matrix estimated on a proxy dataset on the server, to reduce the weight divergence between the local and global model in FL, and thus improve the generalization ability of federated models.

Experiments on FL settings with different local identicalness show that with additional bandwidth reduction tricks, the proposed FedCL can greatly improve the initial accuracy of federated models with as low as 5\% extra communication costs.

\section{Acknowledgement}
This work is supported by the National Natural Science Foundation of China under Grant 61936011 and 61521002, as well as the Beijing Key Lab of Networked Multimedia (Z161100005016051).

\small{
\bibliographystyle{IEEEbib}
\bibliography{refs}

\begin{thebibliography}{10}

\bibitem{konevcny2015federated}
Jakub Kone{\v{c}}n{\`y}, Brendan McMahan, and Daniel Ramage,
\newblock ``Federated optimization: Distributed optimization beyond the
  datacenter,''
\newblock {\em arXiv preprint arXiv:1511.03575}, 2015.

\bibitem{mcmahan2017communication}
Brendan McMahan, Eider Moore, Daniel Ramage, Seth Hampson, and Blaise~Aguera
  y~Arcas,
\newblock ``Communication-efficient learning of deep networks from
  decentralized data,''
\newblock in {\em Artificial Intelligence and Statistics}, 2017, pp.
  1273--1282.

\bibitem{yao2018twostream}
Xin Yao, Chaofeng Huang, and Lifeng Sun,
\newblock ``Two-stream federated learning: Reduce the communication costs,''
\newblock in {\em Visual Communications and Image Processing (VCIP), 2018}.
  IEEE, 2018, pp. 1--4.

\bibitem{yao2019towards}
Xin Yao, Tianchi Huang, Chenglei Wu, Ruixiao Zhang, and Lifeng Sun,
\newblock ``Towards faster and better federated learning: A feature fusion
  approach,''
\newblock in {\em IEEE International Conference on Image Processing}, 2019.

\bibitem{agarwal2018cpsgd}
Naman Agarwal, Ananda~Theertha Suresh, Felix Xinnan~X Yu, Sanjiv Kumar, and
  Brendan McMahan,
\newblock ``cpsgd: Communication-efficient and differentially-private
  distributed sgd,''
\newblock in {\em Advances in Neural Information Processing Systems}, 2018, pp.
  7564--7575.

\bibitem{bonawitz2017practical}
Keith Bonawitz, Vladimir Ivanov, Ben Kreuter, Antonio Marcedone, H~Brendan
  McMahan, Sarvar Patel, Daniel Ramage, Aaron Segal, and Karn Seth,
\newblock ``Practical secure aggregation for privacy-preserving machine
  learning,''
\newblock in {\em Proceedings of the 2017 ACM SIGSAC Conference on Computer and
  Communications Security}. ACM, 2017, pp. 1175--1191.

\bibitem{liu2018secure}
Yang Liu, Tianjian Chen, and Qiang Yang,
\newblock ``Secure federated transfer learning,''
\newblock {\em arXiv preprint arXiv:1812.03337}, 2018.

\bibitem{zhao2018federated}
Yue Zhao, Meng Li, Liangzhen Lai, Naveen Suda, Damon Civin, and Vikas Chandra,
\newblock ``Federated learning with non-iid data,''
\newblock {\em arXiv preprint arXiv:1806.00582}, 2018.

\bibitem{jeong2018communication}
Eunjeong Jeong, Seungeun Oh, Hyesung Kim, Jihong Park, Mehdi Bennis, and
  Seong-Lyun Kim,
\newblock ``Communication-efficient on-device machine learning: Federated
  distillation and augmentation under non-iid private data,''
\newblock {\em arXiv preprint arXiv:1811.11479}, 2018.

\bibitem{yao2019fedmeta}
Xin Yao, Tianchi Huang, Rui-Xiao Zhang, Ruiyu Li, and Lifeng Sun,
\newblock ``Federated learning with unbiased gradient aggregation and
  controllable meta updating,''
\newblock {\em arXiv preprint arXiv:1910.08234}, 2019.

\bibitem{shoham2019overcoming}
Neta Shoham, Tomer Avidor, Aviv Keren, Nadav Israel, Daniel Benditkis, Liron
  Mor-Yosef, and Itai Zeitak,
\newblock ``Overcoming forgetting in federated learning on non-iid data,''
\newblock {\em arXiv preprint arXiv:1910.07796}, 2019.

\bibitem{kirkpatrick2017overcoming}
James Kirkpatrick, Razvan Pascanu, Neil Rabinowitz, Joel Veness, Guillaume
  Desjardins, Andrei~A Rusu, Kieran Milan, John Quan, Tiago Ramalho, Agnieszka
  Grabska-Barwinska, et~al.,
\newblock ``Overcoming catastrophic forgetting in neural networks,''
\newblock {\em Proceedings of the national academy of sciences}, vol. 114, no.
  13, pp. 3521--3526, 2017.

\bibitem{goodfellow2013empirical}
Ian~J Goodfellow, Mehdi Mirza, Da~Xiao, Aaron Courville, and Yoshua Bengio,
\newblock ``An empirical investigation of catastrophic forgetting in
  gradient-based neural networks,''
\newblock {\em arXiv preprint arXiv:1312.6211}, 2013.

\bibitem{li2017learning}
Zhizhong Li and Derek Hoiem,
\newblock ``Learning without forgetting,''
\newblock {\em IEEE transactions on pattern analysis and machine intelligence},
  vol. 40, no. 12, pp. 2935--2947, 2017.

\bibitem{yao2019adversarial}
Xin Yao, Tianchi Huang, Chenglei Wu, Rui-Xiao Zhang, and Lifeng Sun,
\newblock ``Adversarial feature alignment: Avoid catastrophic forgetting in
  incremental task lifelong learning,''
\newblock {\em Neural computation}, vol. 31, no. 11, pp. 2266--2291, 2019.

\bibitem{aljundi2018memory}
Rahaf Aljundi, Francesca Babiloni, Mohamed Elhoseiny, Marcus Rohrbach, and
  Tinne Tuytelaars,
\newblock ``Memory aware synapses: Learning what (not) to forget,''
\newblock in {\em Proceedings of the European Conference on Computer Vision
  (ECCV)}, 2018, pp. 139--154.

\bibitem{zenke2017continual}
Friedemann Zenke, Ben Poole, and Surya Ganguli,
\newblock ``Continual learning through synaptic intelligence,''
\newblock in {\em International Conference on Machine Learning}, 2017, pp.
  3987--3995.

\bibitem{lecun1998gradient}
Yann LeCun, L{\'e}on Bottou, Yoshua Bengio, and Patrick Haffner,
\newblock ``Gradient-based learning applied to document recognition,''
\newblock {\em Proceedings of the IEEE}, vol. 86, no. 11, pp. 2278--2324, 1998.

\bibitem{krizhevsky2009learning}
Alex Krizhevsky and Geoffrey Hinton,
\newblock ``Learning multiple layers of features from tiny images,''
\newblock Tech. {R}ep., Citeseer, 2009.

\bibitem{hsu2019measuring}
Tzu-Ming~Harry Hsu, Hang Qi, and Matthew Brown,
\newblock ``Measuring the effects of non-identical data distribution for
  federated visual classification,''
\newblock {\em arXiv preprint arXiv:1909.06335}, 2019.

\bibitem{li2018federated}
Tian Li, Anit~Kumar Sahu, Manzil Zaheer, Maziar Sanjabi, Ameet Talwalkar, and
  Virginia Smith,
\newblock ``Federated optimization in heterogeneous networks,''
\newblock {\em arXiv preprint arXiv:1812.06127}, 2018.

\end{thebibliography}
}

\end{document}